\documentclass[french,a4paper]{article} 
 \usepackage[utf8]{inputenc}
 \usepackage[T1]{fontenc}
 \usepackage{lmodern} 
 \usepackage{xspace} 
 \usepackage[np]{numprint} 
 \usepackage[main=french]{babel}
 \usepackage{graphicx}
 \title{Lev\'ee d'am\-bi\-gu\"{\i}\-t\'es par grammaires locales}
\author{\'Eric Laporte\thanks{Institut Gaspard-Monge, Universit\'e de Marne-la-Vall\'ee, 2, rue de la Butte-Verte, F-93166 Noisy-le-Grand CEDEX, France.}}
\date{}
\begin{document}
\maketitle

\begin{sloppypar}
\section{Introduction}
\end{sloppypar}

De nombreux mots sont ambigus quant \`a leurs cat\'egories grammaticales~: ainsi, {\it montre} peut \^etre nom ou verbe. Toutefois, lorsqu'un mot appara\^{\i}t dans un texte, cette ambigu\"{\i}t\'e se r\'eduit g\'en\'eralement beaucoup~: dans {\it Une des \'etudes les plus importantes montre que l'exp\'erience est critique}, le mot {\it montre} ne peut \^etre qu'un verbe. Un sys\-t\`eme d'\'eti\-que\-tage lexical est un sys\-t\`eme qui attribue des cat\'egories lexicales aux mots. Lever des ambigu\"{\i}t\'es de cat\'egories lexicales consiste \`a utiliser le contexte pour r\'eduire le nombre de cat\'egories lexicales associ\'ees aux mots. La le\-v\'ee des ambigu\"{\i}t\'es de cat\'egories lexicales est un des principaux d\'efis de l'\'eti\-que\-tage lexical.

Le probl\`eme d'\'etiqueter les mots par des cat\'egories lexicales se pose fr\'equemment dans le traitement des langues naturelles, par exemple pour la correction orthographique, la v\'e\-ri\-fi\-ca\-tion grammaticale ou stylistique, la reconnaissance d'expressions, la phon\'etisation, l'analyse de corpus de textes... En analyse syntaxique, l'\'eti\-que\-tage correct des mots fait partie des r\'esultats de l'analyse, mais si on a pr\'ealablement affect\'e les mots de leurs cat\'egories lexicales, le reste de l'analyse en est souvent facilit\'e (Milne, 1986~; Hindle, 1989~; Rimon, Herz, 1991~; Cutting et al., 1992). Les grands corpus d\'esambigu\"{\i}s\'es sont une vaste source d'informations utiles dans de nombreuses applications, et sont notamment mis \`a contribution pour l'apprentissage des sys\-t\`emes probabilistes, mais leur \'etiquetage manuel est lent, co\^uteux et source d'erreurs. Les sys\-t\`emes d'\'eti\-que\-tage lexical sont ainsi utiles comme composant initial de nombreux sys\-t\`emes de traitement de langues naturelles.

\begin{sloppypar}
\section{M\'ethodologie}
\subsection{Les \'eti\-quettes lexicales}
\end{sloppypar}

Un sys\-t\`eme d'\'eti\-que\-tage lexical attribue \`a chaque forme d'un texte une ou plusieurs \'eti\-quettes, c'est-\`a-dire des codes qui renferment des informations lexicales. Si ces informations se r\'eduisent \`a la cat\'egorie grammaticale, on compte de 10 \`a 20 cat\'egories lexicales. Si elles incluent d'autres donn\'ees grammaticales, les cat\'egories lexicales sont plus fines et plus nombreuses. Ces donn\'ees grammaticales peuvent comporter~:

\begin{itemize}
\item la forme canonique, par exemple {\it montrer} pour {\it montr\'ee}, ou les informations requises pour la reconstituer \`a partir de la forme trouv\'ee dans le texte~;
\item les traits flexionnels~: temps, personne, genre, nombre...~;
\item la d\'elimitation des mots compos\'es, c'est-\`a-dire des s\'equences fig\'ees comportant plusieurs mots simples s\'epar\'es par des s\'eparateurs graphiques, comme {\it bien s\^ur} ou {\it traitement de texte}. Dans ce cas, les compos\'es sont \'etiquet\'es en tant que tels, par exemple {\it Adverbe} pour {\it bien s\^ur}.
\item Si les \'eti\-quettes donnent explicitement les formes canoniques, et aussi les codes de flexion (les num\'eros de conjugaison par exemple), toute forme peut \^etre d\'eduite de son \'eti\-quette.
\end{itemize}
Dans la le\-v\'ee d'am\-bi\-gu\"{\i}\-t\'es lexicales, on prend rarement en compte les informations de plus haut niveau, telles que la relation syntaxique avec le pr\'edicat (Koskenniemi, 1990).

Le corpus \'etiquet\'e Brown utilise un ensemble de 87 \'eti\-quettes simples (Garside, Leech, Sampson, 1987, pp.~165-183) r\'eutilis\'e dans d'autres projets. Pour le fran\c{c}ais, un ensemble d'\'eti\-quettes qui donne seulement la cat\'egorie grammaticale et les traits flexionnels a \`a peu pr\`es la m\^eme taille. Dans cet article, nous d\'ecrivons des exp\'eriences en fran\c{c}ais avec des cat\'egories lexicales qui incluent la forme canonique dans les informations lexicales, \`a la fois pour les mots simples et pour les mots compos\'es. La taille de l'ensemble d'\'eti\-quettes est ainsi celle d'un dictionnaire de la langue.

\begin{sloppypar}
\subsection{La forme du texte d\'esambigu\"{\i}s\'e}
\end{sloppypar}

La plupart des sys\-t\`emes de le\-v\'ee d'am\-bi\-gu\"{\i}\-t\'es produisent pour un texte donn\'e une s\'e\-quence de paires mot/\'eti\-quette~:  chaque mot re\c{c}oit une \'eti\-quette unique. Ce choix peut a priori s'appuyer sur deux pr\'esuppos\'es~:
\begin{enumerate}
\item qu'il est possible de mettre au point un sys\-t\`eme d'\'eti\-que\-tage lexical qui affecte une \'eti\-quette unique \`a chaque mot sans aucune erreur~;
\item ou que le fait d'attribuer \`a un mot dans un texte une seule \'eti\-quette fausse n'est pas un d\'efaut s\'erieux d'un sys\-t\`eme d'\'eti\-que\-tage lexical.
\end{enumerate}
L'affirmation~1 n'a pas un statut clair, sauf \`a consid\'erer un analyseur syntaxique comme un module d'un sys\-t\`eme d'\'eti\-que\-tage lexical au lieu de l'inverse~: m\^eme si l'ensemble d'\'eti\-quettes est rudimentaire, l'\'eti\-que\-tage lexical correct de certaines phrases naturelles met en jeu la reconnaissance de leur structure syntaxique globale ou m\^eme la compr\'ehension de leur sens. L'affirmation~2 est \'egalement contestable, surtout dans le contexte de l'analyse syntaxique~: il est parfois impossible de corriger \`a la main les sorties d'un sys\-t\`eme d'\'eti\-que\-tage lexical~; or il est naturel qu'un analyseur syntaxique \'elimine des hypoth\`eses, il l'est moins qu'il en cr\'ee de nouvelles avec des cat\'egories grammaticales diff\'erentes de celles de d\'epart. De plus, si on \'eti\-quette chaque mot d'une fa\c{c}on unique, m\^eme les phrases effectivement ambigu\"es sont repr\'esent\'ees comme non ambigu\"es.

Un certain nombre de sys\-t\`emes r\'ecents d'\'eti\-que\-tage lexical (Silberztein, 1989~; Koskenniemi, 1990~; Rimon, Herz, 1991~; Roche, 1992) produisent, au contraire, plusieurs solutions lorsque le texte est lexicalement ambigu ou que
l'unique solution correcte ne peut pas \^etre trouv\'ee. Ces contributions visent \`a garantir un taux de silence nul~: la ou les \'eti\-quettes correctes pour un mot ne doivent jamais \^etre \'elimin\'ees. Cet objectif est rarement \'evoqu\'e en-dehors de ces auteurs, et peu r\'ealiste pour les sys\-t\`emes qui \'eti\-quettent chaque mot de fa\c{c}on unique, \`a moins de postuler le pr\'esuppos\'e~1.

\begin{figure}
\centering
\includegraphics[width=11.5cm]{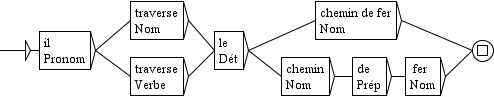}
\caption{un automate acyclique pour {\it Il traverse le chemin de fer.}}
\label{f1}
\end{figure}

Les \'eti\-quettes des mots d'une m\^eme phrase ne sont pas in\-d\'e\-pen\-dantes, c'est pourquoi le r\'esultat d'un sys\-t\`eme \`a plusieurs solutions pour une s\'e\-quence donn\'ee de mots est un ensemble d'une ou plusieurs s\'equences d'\'eti\-quettes. L'ensemble des s\'equences d'\'eti\-quettes s\'electionn\'ees pour une s\'e\-quence d'en\-tr\'ee donn\'ee est repr\'esent\'e sous une forme appropri\'ee. Comme il s'agit d'un ensemble fini de s\'equences qui ont g\'en\'eralement beaucoup en commun, cette forme est toujours celle d'un automate fini acyclique, \'egalement appel\'e graphe orient\'e acyclique (DAG ou DAWG en anglais), machine \`a \'etats finis ou r\'eseau \`a \'etats finis (Koskenniemi, 1990), graphe de phrase (Rimon, Herz, 1991) ou treillis de mots (Vosse, 1992). Un des avantages des automates acycliques dans ce contexte est qu'ils permettent de syst\'ematiser la repr\'esentation des ambigu\"{\i}t\'es lexicales. Ils sont en effet utilisables avant comme apr\`es la le\-v\'ee des ambigu\"{\i}t\'es, et que celles-ci soient li\'ees aux cat\'egories grammaticales, aux traits flexionnels, ou \`a la d\'elimitation des compos\'es (pour repr\'esenter la distinction entre un compos\'e et une s\'e\-quence de mots simples). La figure \ref{f1} illustre l'am\-bi\-gu\"{\i}\-t\'e entre cat\'egories grammaticales pour {\it traverse} et l'am\-bi\-gu\"{\i}\-t\'e entre compos\'e et mots simples pour {\it chemin de fer}.

\begin{sloppypar}
\subsection{\'Etiquetage initial et le\-v\'ee d'am\-bi\-gu\"{\i}\-t\'es}
\end{sloppypar}

La plupart des sys\-t\`emes d'\'eti\-que\-tage lexical divisent la t\^ache en deux \'etapes~: en premier lieu, lors d'un \'etiquetage initial, les formes sont consid\'er\'ees ind\'ependamment de leur contexte pour dresser la liste de toutes les \'eti\-quettes pour chaque mot~; ensuite, lors de la le\-v\'ee d'am\-bi\-gu\"{\i}\-t\'es, on prend en compte le contexte pour s\'electionner une partie de l'ensemble des s\'equences \'etiquet\'ees initiales.

Dans d'autres sys\-t\`emes d'\'eti\-que\-tage lexical (Klein, Simmons, 1963~; Dermatas, Kokkinakis, 1989~; Pelillo, Refice, 1991~; Brill, 1992~; Federici, Pirrelli, 1992), les deux sous-t\^aches sont effectu\'ees en m\^eme temps et un r\'esultat d\'esambigu\"{\i}s\'e est construit directement, g\'en\'eralement pour \'eviter la construction d'un grand dictionnaire.

Plusieurs arguments viennent en faveur de la solution modulaire. Les deux sous-t\^aches sont clairement d\'efinies. Une fois choisi un ensemble d'\'eti\-quettes lexicales, les deux sous-t\^aches sont in\-d\'e\-pen\-dantes. Il peut en \^etre de m\^eme des m\'ethodes permettant de les effectuer avec les meilleurs r\'esultats~: perfectionner l'\'eti\-que\-tage initial est un probl\`eme de description morphologique des mots, am\'eliorer la le\-v\'ee des ambigu\"{\i}t\'es met en jeu la description grammaticale de s\'equences de mots.

Cette solution est coh\'erente avec l'utilisation d'automates acycliques pour repr\'esenter les ambigu\"{\i}t\'es lexicales. Apr\`es l'\'eti\-que\-tage initial, l'ensemble des s\'equences reconnues par l'automate est l'ensemble des s\'equences d'\'eti\-quettes possibles a priori pour la s\'e\-quence d'en\-tr\'ee. Lors de la le\-v\'ee d'am\-bi\-gu\"{\i}\-t\'es, l'automate est modifi\'e. Le nombre de s\'equences reconnues par l'automate diminue, mais le nombre d'\'etats et de transitions dans l'automate peut cro\^{\i}tre ou d\'ecro\^{\i}tre.

La solution modulaire a bien s\^ur un int\'er\^et particulier si l'on dispose d'un dictionnaire morphologique fiable qui donne pour chaque forme, simple ou compos\'ee, la liste des \'eti\-quettes possibles. Dans ce cas, l'\'eti\-que\-tage initial est simplement men\'e \`a bien par une consultation du dictionnaire. Un tel environnement pour le fran\c{c}ais a \'et\'e d\'evelopp\'e au LADL\footnote{Laboratoire d'automatique documentaire et linguistique, Universit\'e Paris~7, 2, place Jussieu, F-75252 Paris CEDEX 05, France.} et au CERIL\footnote{Centre d'\'etudes et de recherches en informatique linguistique, Institut Gaspard-Monge.} avec les dictionnaires DELAF (Courtois, 1990) et DELACF (Silberztein, 1990), et avec les algorithmes de compression et de consultation mis en \oe uvre par Revuz (1991) et Roche (1992) pour obtenir de meilleures performances qu'avec les arbres lexicographiques ({\it tries}) de Knuth (1973). Il est maintenant int\'egr\'e au sys\-t\`eme d'analyse lexicale {\tt INTEX} (Silberztein, 1993). La taille du dictionnaire comprim\'e est inf\'erieure \`a 900~Ko pour 700.000 formes.

\begin{sloppypar}
\subsection{Donn\'ees \'elabor\'ees \`a la main ou donn\'ees statistiques}
\end{sloppypar}

Les informations utilis\'ees par les sys\-t\`emes d'\'eti\-que\-tage lexical pour lever des ambigu\"{\i}t\'es consistent soit en connaissances grammaticales formalis\'ees \`a la main, soit en donn\'ees statistiques acquises par apprentissage automatique dans un grand corpus de textes. Le sys\-t\`eme de Hindle (1989) utilise un m\'elange des deux. Comme exemples de sys\-t\`emes d'\'eti\-que\-tage lexical qui utilisent des donn\'ees grammaticales \'elabor\'ees \`a la main, on peut citer ceux de Klein, Simmons (1963), Hindle (1983), Silberztein (1989), Paulussen, Martin (1992), Roche (1992). Dans le cas de Rimon, Herz (1991), les donn\'ees sont produites automatiquement \`a partir de grammaires alg\'ebriques faites \`a la main. Dans toutes ces contributions, les donn\'ees grammaticales sont formalis\'ees dans des automates finis, ou pourraient facilement l'\^etre. Nous d\'esignons ces donn\'ees par le nom de grammaires locales, car elles ne constituent jamais une grammaire compl\`ete de la langue. L'\'eti\-que\-tage lexical \`a partir de statistiques est repr\'esent\'e par Greene, Rubin (1971), Hindle (1989), Brill (1992), Federici, Pirrelli (1992), qui utilisent des r\`egles produites par des processus statistiques~; et par Marshall (1983), Jelinek (1985), DeRose (1988), Cutting et al. (1992), etc., qui utilisent des tables de statistiques, par exemple des param\`etres de mod\`eles de Markov. Toutes ces contributions donnent \`a chaque mot une \'eti\-quette unique.

Les arguments pour ou contre ces deux types de m\'ethodes tiennent souvent \`a leur capacit\'e \`a faire face \`a la vari\'et\'e qui caract\'erise le texte r\'eel dans toute sa g\'en\'eralit\'e. Certains doutent que des connaissances linguistiques \'elabor\'ees \`a la main puissent tenir compte de tout ce qui peut se pr\'esenter dans du texte. D'autres pensent que des informations apprises automatiquement dans un corpus, m\^eme vaste et soigneusement compos\'e d'\'echantillons vari\'es de types de textes, ne sont pas assez pr\'ecises pour de nouveaux textes. L'utilisation de connaissances linguistiques \'elabor\'ees \`a la main nous semble en tous cas mieux adapt\'ee pour atteindre l'objectif d'un taux de silence nul, c'est-\`a-dire pour garantir qu'une analyse n'est \'elimin\'ee que si elle est sans aucun doute incorrecte. L'auteur des donn\'ees linguistiques peut s'aider d'un corpus de textes, mais doit pouvoir cr\'eer des contre-exemples qui n'y figurent pas, de sorte que les donn\'ees linguistiques qu'il \'elabore soient in\-d\'e\-pen\-dantes de ce corpus.

La suite de cet article concerne une m\'ethode de le\-v\'ee d'am\-bi\-gu\"{\i}\-t\'es lexicales adapt\'ee \`a l'objectif d'un taux de silence nul (Silberztein, 1989, 1993) et mise en \oe uvre dans le sys\-t\`eme {\tt INTEX} de Silberztein (1993). Nous pr\'esentons ici une description formelle de cette m\'ethode. Elle combine la possibilit\'e de plusieurs solutions --- le r\'esultat produit pour une s\'e\-quence de mots donn\'ee est un automate acyclique ---, le parti pris modulaire --- l'\'eti\-que\-tage lexical et la le\-v\'ee des ambigu\"{\i}t\'es sont consid\'er\'es comme in\-d\'e\-pen\-dants une fois choisi un ensemble d'\'eti\-quettes ---, et l'utilisation de connaissances linguistiques \'elabor\'ees \`a la main, pour l'\'eti\-que\-tage initial --- de grands dictionnaires morphologiques --- comme pour la le\-v\'ee des ambigu\"{\i}t\'es --- des donn\'ees grammaticales appel\'ees grammaires locales. La seule autre contribution qui rentre dans ce cadre est, \`a notre connaissance, celle de Roche (1992). Toutes les deux comprennent des algorithmes et leur mise en \oe uvre et elles utilisent les m\^emes dictionnaires. Pour une comparaison entre ces deux sys\-t\`emes, cf. Laporte (1994).

\begin{sloppypar}
\section{Les \'eti\-quettes lexicales dans INTEX}
\subsection{\'Etiquettes grammaticales compl\`etes}
\end{sloppypar}

La confrontation d'un texte avec les dictionnaires produit plusieurs s\'equences d'\'eti\-quettes grammaticales en raison des ambigu\"{\i}t\'es lexicales. Ainsi, pour la phrase {\it Je ne me le suis pas fait confirmer sur le moment}, on obtient l'\'eti\-que\-tage initial suivant~:

\begin{quotation}
(''{\it je.}'' + ''{\it je.PRO:1s}'')

(''{\it ne.}'' + ''{\it ne.XI}'' + ''{\it ne.XI[+ Pr\'ed]}'')

''{\it me.PRO:1s}''

(''{\it le.DET:ms}'' + ''{\it le.PRO:3ms}'')

(''{\it \^etre.V:P1s}'' + ''{\it suivre.V:P1s:P2s:Y2s}'')

(''{\it pas.ADV}'' + ''{\it pas.N:ms:mp}'' + ''{\it pas.XI}'')

(''{\it faire.V:Kms:P3s}'' + ''{\it fait.A:ms}'' + ''{\it fait.N:ms}'' + ''{\it fait.XI[+ Pr\'ed]}'')

''{\it confirmer.V:W}''

(

''{\it sur/le/moment.ADV;PDETC}''

+

(''{\it sur.A:ms}'' + ''{\it sur.PREP}'')

(''{\it le.DET:ms}'' + ''{\it le.PRO:3ms}'')

''{\it moment.N:ms}''

)
\end{quotation}
Les unit\'es de base d'{\tt INTEX} donnent la forme canonique (mot simple ou mot compos\'e), la cat\'egorie grammaticale, et les traits flexionnels s'ils sont pertinents. Nous les appellerons \'eti\-quettes grammaticales compl\`etes et nous les noterons comme dans les exemples suivants~: \(<\){\it suivre V:P2s}\(>\), \(<\){\it sur/le/moment ADV;PDETC}\(>\), \(<\){\it coup/fumant N;NA:ms}\(>\)...

\begin{sloppypar}
\subsection{\'Etiquettes grammaticales incompl\`etes}
\end{sloppypar}

Le sys\-t\`eme de le\-v\'ee d'am\-bi\-gu\"{\i}\-t\'es grammaticales d'{\tt INTEX} utilise un autre ensemble d'\'eti\-quettes lexicales que nous appellerons \'eti\-quettes grammaticales incompl\`etes et qui peuvent sp\'ecifier

\begin{itemize}
\item une forme canonique~: \(<\){\it  prendre}\(>\), \(<\){\it  le}\(>\), \(<\){\it  coup/fumant}\(>\)...
\item une cat\'egorie grammaticale~: \(<\){\it  V}\(>\), \(<\){\it  ADV}\(>\), \(<\){\it  A}\(>\)...
\begin{sloppypar}
\item une forme canonique et des traits flexionnels pertinents pour cette forme~: \(<\){\it  prendre:P3s}\(>\), \(<\){\it  prendre:P}\(>\), \(<\){\it  prendre:s}\(>\), \(<\){\it  coup/fumant:ms}\(>\)...
\end{sloppypar}
\item une cat\'egorie grammaticale et des traits flexionnels pertinents pour cette cat\'egorie~: \(<\){\it  V:P3s}\(>\), \(<\){\it  V:P}\(>\), \(<\){\it  V:s}\(>\), \(<\){\it  N:ms}\(>\)...
\item une forme simple~: {\it vient}, {\it aussit\^ot}, {\it fait}...
\item L'\'eti\-quette \(<\){\it  MOT}\(>\) repr\'esente toutes les formes simples, \`a l'exclusion des formes compos\'ees.
\end{itemize}
Une \'eti\-quette grammaticale incompl\`ete sert \`a repr\'esenter l'ensemble des \'eti\-quettes grammaticales compl\`etes qui satisfont aux informations qu'elle sp\'ecifie. Exemples~:
\begin{quote}
\(<\){\it  prendre}\(>\) repr\'esente toutes les formes du verbe {\it prendre}

\(<\){\it  A}\(>\) repr\'esente tous les adjectifs \`a toutes les formes

{\it  suis} repr\'esente toutes les \'eti\-quettes lexicales possibles de {\it  suis}, c'est-\`a-dire une forme du verbe {\it  \^etre} et trois formes du verbe {\it  suivre}.
\end{quote}
En d'autres termes, une \'eti\-quette grammaticale compl\`ete est ou n'est pas conforme \`a une \'eti\-quette grammaticale incompl\`ete. Exemples~:
\begin{quote}
\(<\){\it  suivre V:P2s}\(>\) est conforme \`a \(<\){\it  suivre}\(>\) et \`a \(<\){\it  V}\(>\)

\(<\){\it  \^etre V:P1s}\(>\) est conforme \`a \(<\){\it  \^etre}\(>\) et \`a \(<\){\it  V:P}\(>\)

\(<\){\it  suivre V:P2s}\(>\) n'est pas conforme \`a \(<\){\it  \^etre}\(>\) car la forme canonique ne correspond pas

\(<\){\it  coup/fumant N;NA:ms}\(>\) n'est pas conforme \`a \(<\){\it  MOT}\(>\) car c'est une forme compos\'ee

\(<\){\it  suivre V:P1s}\(>\) et \(<\){\it  suivre V:P2s}\(>\) sont conformes \`a {\it suis}

\(<\){\it  \^etre V:P1s}\(>\) est conforme \`a {\it suis}
\end{quote}

\begin{sloppypar}
\section{Grammaires locales de le\-v\'ee d'am\-bi\-gu\"{\i}\-t\'es}
\end{sloppypar}

Dans le sys\-t\`eme {\tt INTEX}, la le\-v\'ee d'am\-bi\-gu\"{\i}\-t\'es lexicales met en jeu des informations linguistiques \'elabor\'ees \`a la main. Ces informations sont sp\'ecifi\'ees sous la forme d'un ou plusieurs transducteurs appel\'es grammaires locales de le\-v\'ee d'am\-bi\-gu\"{\i}\-t\'es. On peut voir une telle grammaire locale comme un moyen formel de sp\'ecifier un ensemble de s\'equences grammaticales accept\'ees. Un algorithme de le\-v\'ee d'am\-bi\-gu\"{\i}\-t\'es applique la grammaire \`a un texte, ce qui consiste \`a s\'electionner parmi les \'etiquetages grammaticaux du texte ceux qui sont conformes \`a la grammaire. L'int\'er\^et de ce sys\-t\`eme est qu'il permet de sp\'ecifier des ensembles de s\'equences grammaticales tout \`a fait complexes \`a l'aide de grammaires relativement petites, lisibles et intuitives. Il est important que les grammaires ne rejettent jamais une s\'e\-quence grammaticale correcte. En revanche, on est loin de disposer d'une grammaire qui n'accepte que les s\'equences correctes.

\begin{figure}
\centering
\includegraphics[width=11.5cm]{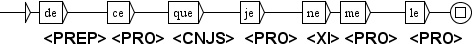}
\caption{le transducteur \(T_1\).}
\label{graphet1}
\end{figure}

Nous utilisons l'\'editeur d'automates {\tt Editor} de Max Silberztein pour repr\'esenter graphiquement les transducteurs de le\-v\'ee d'am\-bi\-gu\"{\i}\-t\'es (exemple~: le transducteur \(T_1\), figure~\ref{graphet1}). Chaque transition comporte deux \'eti\-quettes grammaticales incompl\`etes, une dite ``d'en\-tr\'ee'' (dans la bo\^{\i}te) et une ``de sortie'' (sous la bo\^{\i}te). Dans un transducteur, une s\'e\-quence d'en\-tr\'ee est une s\'e\-quence d'\'eti\-quettes grammaticales d'en\-tr\'ee. Le transducteur sert \`a mettre en relation des s\'equences d'en\-tr\'ee et des s\'equences de sortie. Par exemple, le transducteur \(T_1\) (figure~\ref{graphet1}) met en relation la s\'e\-quence d'en\-tr\'ee
\begin{quote}
{\it de ce que je ne me le}
\end{quote}
et la s\'e\-quence de sortie
\begin{quote}
\(<\){\it  PREP}\(>\) \(<\){\it  PRO}\(>\) \(<\){\it  CNJS}\(>\) \(<\){\it  PRO}\(>\) \(<\){\it  XI}\(>\) \(<\){\it  PRO}\(>\) \(<\){\it  PRO}\(>\)
\end{quote}
Grosso modo, les s\'equences d'en\-tr\'ee servent \`a s\'electionner les portions du texte auxquelles la grammaire locale va s'appliquer, et les s\'equences de sortie imposent des contraintes \`a ces portions de texte. Pour s'assurer qu'une grammaire ne rejettera pas de s\'equences grammaticales correctes, on a besoin d'une r\`egle plus pr\'ecise pour r\'epondre \`a la question~: \'etant donn\'e une s\'e\-quence grammaticale compl\`ete et une grammaire locale de le\-v\'ee d'am\-bi\-gu\"{\i}\-t\'es, la grammaire accepte-t-elle la s\'equence~?

\begin{sloppypar}
\subsection{Cas o\`u les s\'equences d'en\-tr\'ee du transducteur ne sont constitu\'ees que de formes simples}
\end{sloppypar}

Dans ce premier cas particulier, on a une r\`egle plus simple que dans le cas g\'en\'eral. Nous illustrerons cette r\`egle en prenant l'exemple du transducteur \(T_1\) (figure~\ref{graphet1}) et de la s\'e\-quence grammaticale compl\`ete
\begin{quote}
\(<\){\it  cela PRO:ms}\(>\) \(<\){\it  venir V:P3s}\(>\) \(<\){\it  de PREP}\(>\) \(<\){\it  ce PRO:3s}\(>\) \(<\){\it  que CNJS}\(>\) \(<\){\it  je PRO:1s}\(>\) \(<\){\it  ne XI}\(>\) \(<\){\it  me PRO:1s}\(>\) \(<\){\it  le PRO:3ms}\(>\) \(<\){\it  \^etre V:P1s}\(>\) \(<\){\it  pas ADV}\(>\) \(<\){\it  faire V:Kms}\(>\) \(<\){\it  confirmer V:W}\(>\) \(<\){\it  aussit\^ot ADV}\(>\)
\end{quote}
Cette s\'e\-quence est l'\'eti\-que\-tage grammatical correct de
\begin{quote}
{\it Cela vient de ce que je ne me le suis pas fait confirmer aussit\^ot}
\end{quote}

La grammaire accepte la s\'e\-quence si et seulement si on peut diviser la s\'e\-quence en portions qui sont chacune de l'un {\em ou} de l'autre des deux types suivants~:
\begin{enumerate}
\item la portion est conforme \`a la fois \`a une s\'e\-quence d'en\-tr\'ee et \`a une s\'e\-quence de sortie associ\'ees par le transducteur.
\item la portion est r\'eduite \`a une seule \'eti\-quette grammaticale compl\`ete, peu importe laquelle, mais le texte qui commence \`a ce mot n'appara\^{\i}t dans aucune s\'e\-quence d'en\-tr\'ee du transducteur.
\end{enumerate}

\begin{table}
\centering
\begin{tabular}{|c|l|c|} \hline
S\'equences d'en\-tr\'ee &	S\'equence gramm. &	S\'equences de sortie \\
du transducteur &	compl\`ete &	du transducteur \\ \hline
-- &	\(<\){\it  cela PRO:ms}\(>\) &	-- \\ \hline
-- &	\(<\){\it  venir V:P3s}\(>\) &	-- \\ \hline
de &	\(<\){\it  de PREP}\(>\) &	\(<\)PREP\(>\) \\
ce &	\(<\){\it  ce PRO:3s}\(>\) &	\(<\)PRO\(>\) \\
que &	\(<\){\it  que CNJS}\(>\) &	\(<\)CNJS\(>\) \\
je &	\(<\){\it  je PRO:1s}\(>\) &	\(<\)PRO\(>\) \\
ne &	\(<\){\it  ne XI}\(>\) &	\(<\)XI\(>\) \\
me &	\(<\){\it  me PRO:1s}\(>\) &	\(<\)PRO\(>\) \\
le &	\(<\){\it  le PRO:3ms}\(>\) &	\(<\)PRO\(>\) \\ \hline
-- &	\(<\){\it  \^etre V:P1s}\(>\) &	-- \\ \hline
-- &	\(<\){\it  pas ADV}\(>\) &	-- \\ \hline
-- &	\(<\){\it  faire V:Kms}\(>\) &	-- \\ \hline
-- &	\(<\){\it  confirmer V:W}\(>\) &	-- \\ \hline
-- &	\(<\){\it  aussit\^ot ADV}\(>\) &	-- \\ \hline
\end{tabular}
\caption{une s\'e\-quence grammaticale accept\'ee par le transducteur \(T_1\).}
\label{t1}
\end{table}

Dans cet exemple (table~\ref{t1}), les deux premi\`eres portions sont du type 2~: \(<\){\it  cela PRO:ms}\(>\) et \(<\){\it  venir V:P3s}\(>\). La suivante est du type 1~:
\begin{quote}
\(<\){\it  de PREP}\(>\) \(<\){\it  ce PRO:3s}\(>\) \(<\){\it  que CNJS}\(>\) \(<\){\it  je PRO:1s}\(>\) \(<\){\it  ne XI}\(>\) \(<\){\it  me PRO:1s}\(>\) \(<\){\it  le PRO:3ms}\(>\)
\end{quote}
Toutes les autres sont du type 2. Finalement cette s\'e\-quence est accept\'ee par la grammaire.

On notera que la condition~2 ci-dessus met en jeu non pas seulement l'\'eti\-quette grammaticale compl\`ete qui figure dans la portion consid\'er\'ee, mais aussi les autres \'eti\-quettes compl\`etes associables au m\^eme mot, et \'eventuellement les \'eti\-quettes des mots qui le suivent. L'\'ecriture de grammaires locales de le\-v\'ee d'am\-bi\-gu\"{\i}\-t\'es s'apparente \`a la programmation mais n\'ecessite aussi de l'intuition grammaticale, pour imaginer les s\'equences grammaticales qu'une grammaire va rejeter.

\begin{sloppypar}
\subsection{Deuxi\`eme cas particulier}
\end{sloppypar}

En fait, une r\`egle tr\`es voisine reste valable pour de nombreux autres transducteurs. Il s'agit des transducteurs dont chaque transition v\'erifie l'une des deux conditions suivantes~:
\begin{itemize}
\item l'\'eti\-quette en entr\'ee est une forme simple,
\item ou bien toute \'eti\-quette grammaticale compl\`ete conforme \`a l'\'eti\-quette de sortie est conforme \`a l'\'eti\-quette d'en\-tr\'ee.
\end{itemize}
\begin{figure}
\centering
\includegraphics[width=8cm]{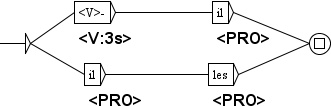}
\caption{le transducteur \(T_2\).}
\label{graphet2}
\end{figure}
Par exemple, c'est le cas pour le  transducteur \(T_2\) (figure~\ref{graphet2}) et pour toutes les grammaires locales de le\-v\'ee d'am\-bi\-gu\"{\i}\-t\'es du chapitre~8 de Silberztein (1993). La r\`egle est alors la suivante.

La grammaire accepte la s\'e\-quence si et seulement si on peut diviser la s\'e\-quence en portions qui sont chacune de l'un {\em ou} de l'autre des deux types suivants~:
\begin{enumerate}
\item la portion est conforme \`a la fois \`a une s\'e\-quence d'en\-tr\'ee et \`a une s\'e\-quence de sortie associ\'ees par le transducteur.
\item la portion est r\'eduite \`a une seule \'eti\-quette grammaticale compl\`ete, peu importe laquelle, mais le texte qui commence \`a ce mot n'admet aucun \'etiquetage grammatical qui corresponde \`a une s\'e\-quence d'en\-tr\'ee du transducteur.
\end{enumerate}
Nous illustrons cette r\`egle avec \(T_2\) et la s\'e\-quence grammaticale
\begin{quote}
\(<\){\it  ne XI[+ Pr\'ed]}\(>\) \(<\){\it  faire V:P3s}\(>\)-\(<\){\it  il PRO:3ms}\(>\) \(<\){\it  le DET:mp}\(>\) \(<\){\it  compte N:mp}\(>\) \(<\){\it  que CNJS}\(>\)
\end{quote}
qui est une partie de l'\'eti\-que\-tage correct de
\begin{quote}
{\it Ne fait-il les comptes que pour rendre service~?}
\end{quote}
Dans cet exemple (table~\ref{t2}), \(<\){\it  il PRO:3ms}\(>\) \(<\){\it  le DET:mp}\(>\) n'est pas compar\'e aux s\'equences d'en\-tr\'ee de la grammaire, car deux portions ne peuvent pas se chevaucher.

\begin{table}
\centering
\begin{tabular}{|c|l|c|} \hline
S\'equences d'en\-tr\'ee &	S\'equence gramm. &	S\'equences de sortie \\
du transducteur &	compl\`ete &	du transducteur \\ \hline
-- &	\(<\){\it  ne XI[+ Pr\'ed]}\(>\) &	-- \\ \hline
\(<\)V\(>\)- &	\(<\){\it  faire V:P3s}\(>\)- &	\(<\)V:3s\(>\) \\
il &	\(<\){\it  il PRO:3ms}\(>\) &	\(<\)PRO\(>\) \\ \hline
-- &	\(<\){\it  le DET:mp}\(>\) &	-- \\ \hline
-- &	\(<\){\it  compte N:mp}\(>\) &	-- \\ \hline
-- &	\(<\){\it  que CNJS}\(>\) &	-- \\ \hline
 &	... &	 \\ \hline
\end{tabular}
\caption{une s\'e\-quence grammaticale accept\'ee par le transducteur \(T_2\).}
\label{t2}
\end{table}

\begin{figure}
\centering
\includegraphics[width=8cm]{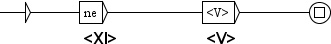}
\caption{le transducteur \(T_3\).}
\label{graphet3}
\end{figure}

Il n'est pas absurde d'avoir des transitions dans lesquelles l'\'eti\-quette d'en\-tr\'ee est la m\^eme que l'\'eti\-quette de sortie. Par exemple, le transducteur \(T_3\) (figure~\ref{graphet3}) accepte la s\'e\-quence ci-dessus mais la rejette si on remplace \(<\){\it  faire V:P3s}\(>\) par \(<\){\it  fait N:ms}\(>\) (table~\ref{t3}).

\begin{table}
\centering
\begin{tabular}{|c|l|c|} \hline
S\'equences d'en\-tr\'ee &	S\'equence gramm. &	S\'equences de sortie \\
du transducteur &	compl\`ete &	du transducteur \\ \hline
ne &	\(<\){\it  ne XI[+ Pr\'ed]}\(>\) &	\(<\)XI\(>\) \\
\(<\)V\(>\) &	\(<\){\it  fait N:ms}\(>\)- &	\(<\)V\(>\) \\ \hline
-- &	\(<\){\it  il PRO:3ms}\(>\) &	-- \\ \hline
-- &	\(<\){\it  le DET:mp}\(>\) &	-- \\ \hline
-- &	\(<\){\it  compte N:mp}\(>\) &	-- \\ \hline
-- &	\(<\){\it  que CNJS}\(>\) &	-- \\ \hline
 &	... &	 \\ \hline
\end{tabular}
\caption{une s\'e\-quence grammaticale rejet\'ee par le transducteur \(T_3\).}
\label{t3}
\end{table}
En effet, la portion \(<\){\it  ne XI[+ Pr\'ed]}\(>\) \(<\){\it  fait N:ms}\(>\) n'est conforme ni \`a la s\'e\-quence d'en\-tr\'ee ni \`a la s\'e\-quence de sortie, alors que le texte {\it ne fait} admet un autre \'etiquetage grammatical conforme \`a la s\'e\-quence d'en\-tr\'ee~: \(<\){\it  ne XI}\(>\) \(<\){\it  faire V:P3s}\(>\). Il n'existe donc aucun d\'ecoupage en portions qui satisfasse aux conditions de la r\`egle ci-dessus.

\begin{sloppypar}
\section{Combinaisons de grammaires locales}
\end{sloppypar}

\begin{figure}
\centering
\includegraphics[width=8cm]{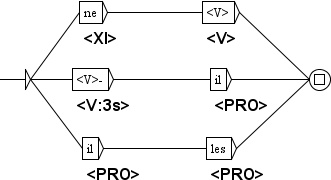}
\caption{le transducteur \(T_2 | T_3\).}
\label{graphet2t3}
\end{figure}

{\tt INTEX} permet d'appliquer plusieurs grammaires locales de le\-v\'ee d'am\-bi\-gu\"{\i}\-t\'es \`a un m\^eme texte. Dans ce cas, tout se passe comme si on avait combin\'e les diff\'erents transducteurs en leur donnant le m\^eme \'etat initial et le m\^eme \'etat final. Par exemple, la combinaison de \(T_2\) (figure~\ref{graphet2}) et de \(T_3\) (figure~\ref{graphet3}), not\'ee \(T_2 | T_3\), est repr\'esent\'ee figure~\ref{graphet2t3}. Les r\`egles ci-dessus s'appliquent alors \`a l'ensemble.

\begin{figure}
\centering
\includegraphics[width=8cm]{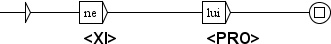}
\caption{le transducteur \(T_4\).}
\label{graphet4}
\end{figure}

Si une grammaire locale rejette une s\'e\-quence grammaticale, et si on combine la grammaire avec une autre, la combinaison obtenue peut accepter la s\'equence. Par exemple, le transducteur \(T_3\) rejette \`a tort la s\'equence
\begin{quote}
\(<\){\it  ne XI}\(>\) \(<\){\it  lui PRO:3s}\(>\) \(<\){\it  dire V:Y2s}\(>\) \(<\){\it  pas ADV}\(>\)
\end{quote}
qui est l'\'eti\-que\-tage correct de {\it Ne lui dis pas}, \`a cause du fait que {\it lui} est aussi une forme du verbe {\it luire}~: \(<\){\it  luire V:Kms}\(>\). Pour tenter de corriger ce d\'efaut, on peut combiner \(T_3\) avec \(T_4\) (figure~\ref{graphet4}). La s\'e\-quence correcte ci-dessus est accept\'ee par \(T_4\) et aussi par la combinaison \(T_3 | T_4\) (table~\ref{t3t4}).

\begin{table}
\centering
\begin{tabular}{|c|l|c|} \hline
S\'equences d'en\-tr\'ee &	S\'equence gramm. &	S\'equences de sortie \\
du transducteur &	compl\`ete &	du transducteur \\ \hline
ne &	\(<\){\it  ne XI}\(>\) &	\(<\)XI\(>\) \\
lui &	\(<\){\it  lui PRO:3s}\(>\) &	\(<\)PRO\(>\) \\ \hline
-- &	\(<\){\it  dire V:Y2s}\(>\) &	-- \\ \hline
-- &	\(<\){\it  pas ADV}\(>\) &	-- \\ \hline
\end{tabular}
\caption{une s\'e\-quence grammaticale accept\'ee par le transducteur \(T_3 | T_4\).}
\label{t3t4}
\end{table}

Inversement, la s\'e\-quence incorrecte
\begin{quote}
\(<\){\it  ne XI}\(>\) \(<\){\it  luire V:Kms}\(>\) \(<\){\it  dire V:Y2s}\(>\) \(<\){\it  pas ADV}\(>\)
\end{quote}
est rejet\'ee par \(T_4\) mais accept\'ee par la combinaison \(T_3 | T_4\).

Si deux grammaires locales acceptent une m\^eme s\'e\-quence grammaticale, la combinaison des deux grammaires peut rejeter la s\'equence. Par exemple, les transducteurs \(T_2\) et \(T_3\) acceptent tous les deux la s\'e\-quence correcte
\begin{quote}
\(<\){\it  ne XI[+ Pr\'ed]}\(>\) \(<\){\it  faire V:P3s}\(>\)-\(<\){\it  il PRO:3ms}\(>\) \(<\){\it  le DET:mp}\(>\) \(<\){\it  compte N:mp}\(>\)
\end{quote}
mais la combinaison \(T_2 | T_3\) (figure~\ref{graphet2t3}) la rejette (table~\ref{t2t3}).

\begin{table}
\centering
\begin{tabular}{|c|l|c|} \hline
S\'equences d'en\-tr\'ee &	S\'equence gramm. &	S\'equences de sortie \\
du transducteur &	compl\`ete &	du transducteur \\ \hline
ne &	\(<\){\it  ne XI[+ Pr\'ed]}\(>\) &	\(<\)XI\(>\) \\
\(<\)V\(>\) &	\(<\){\it  faire V:P3s}\(>\)- &	\(<\)V\(>\) \\ \hline
il &	\(<\){\it  il PRO:3ms}\(>\) &	\(<\)PRO\(>\) \\
les &	\(<\){\it  le DET:mp}\(>\) &	\(<\)PRO\(>\) \\ \hline
-- &	\(<\){\it  compte N:mp}\(>\) &	-- \\ \hline
-- &	\(<\){\it  que CNJS}\(>\) &	-- \\ \hline
 &	... &	 \\ \hline
\end{tabular}
\caption{une s\'e\-quence grammaticale rejet\'ee par le transducteur \(T_2 | T_3\).}
\label{t2t3}
\end{table}

\begin{figure}
\centering
\includegraphics[width=8cm]{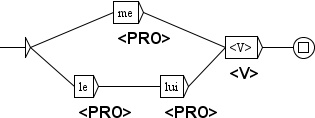}
\caption{le transducteur \(T_5\).}
\label{graphet5}
\end{figure}
\begin{figure}
\centering
\includegraphics[width=8cm]{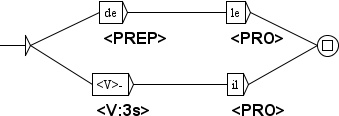}
\caption{le transducteur \(T_6\).}
\label{graphet6}
\end{figure}

Si deux grammaires locales rejettent une m\^eme s\'e\-quence grammaticale, la combinaison des deux grammaires peut accepter la s\'equence. Par exemple, chacun des deux transducteurs  \(T_5\) (figure~\ref{graphet5}) et  \(T_6\) (figure~\ref{graphet6}), utilis\'e s\'epar\'ement, rejette la s\'e\-quence incorrecte
\begin{quote}
\(<\){\it  pourquoi ADV}\(>\) \(<\){\it  me PRO:3s}\(>\) \(<\){\it  presser V:P3p}\(>\)-\(<\){\it  il PRO:3ms}\(>\) \(<\){\it  de PREP}\(>\) \(<\){\it  le PRO:3ms}\(>\) \(<\){\it  luire V:Kms}\(>\) \(<\){\it  dire V:W}\(>\),
\end{quote}
qui est un \'etiquetage du texte incorrect
\begin{quote}
{\it Pourquoi me pressent-il de le lui dire~?}
\end{quote}
Mais la combinaison \(T_5 | T_6\) accepte cette m\^eme s\'e\-quence (table~\ref{t5t6}).
\begin{table}
\centering
\begin{tabular}{|c|l|c|} \hline
S\'equences d'en\-tr\'ee &	S\'equence gramm. &	S\'equences de sortie \\
du transducteur &	compl\`ete &	du transducteur \\ \hline
-- &	\(<\){\it  pourquoi ADV}\(>\) &	-- \\ \hline
me &	\(<\){\it me PRO:3s}\(>\) &	\(<\)PRO\(>\) \\
\(<\)V\(>\) &	\(<\){\it presser V:P3p}\(>\)- &	\(<\)V\(>\) \\ \hline
-- &	\(<\){\it il PRO:3ms}\(>\) &	-- \\ \hline
de &	\(<\){\it de PREP}\(>\) &	\(<\)PREP\(>\) \\
le &	\(<\){\it le PRO:3ms}\(>\) &	\(<\)PRO\(>\) \\ \hline
-- &	\(<\){\it luire V:Kms}\(>\) &	-- \\ \hline
-- &	\(<\){\it dire V:W}\(>\) &	-- \\ \hline
\end{tabular}
\caption{une s\'e\-quence grammaticale accept\'ee par le transducteur \(T_5 | T_6\).}
\label{t5t6}
\end{table}

On voit que pour v\'erifier une grammaire locale de le\-v\'ee d'am\-bi\-gu\"{\i}\-t\'es, il ne suffit pas de consid\'erer les chemins du transducteur s\'epar\'ement~: on a besoin de v\'erifier leurs interactions. De m\^eme, si on utilise une combinaison de plusieurs transducteurs, on ne peut pas pr\'evoir le r\'esultat en les consid\'erant isol\'ement les uns des autres.

\begin{sloppypar}
\section{Cas g\'en\'eral}
\end{sloppypar}

\'Etant donn\'e une s\'e\-quence grammaticale compl\`ete et une grammaire locale de le\-v\'ee d'am\-bi\-gu\"{\i}\-t\'es, la grammaire accepte-t-elle la s\'equence~? Contrairement aux r\`egles donn\'ees plus haut, la r\`egle ci-dessous r\'epond \`a cette question dans le cas g\'en\'eral.

Pour \'enoncer cette r\`egle on a besoin de la notion suivante~: deux s\'equences grammaticales compl\`etes sont \'equivalentes si et seulement si elles d\'ecrivent le m\^eme texte avec la m\^eme d\'elimitation des mots simples et des mots compos\'es. Par exemple,
\begin{quote}
\(<\){\it superbe N:fs}\(>\) \(<\){\it gaulliste A:fs}\(>\)
\end{quote}
et
\begin{quote}
\(<\){\it superbe A:fs}\(>\) \(<\){\it gaulliste N:fs}\(>\)
\end{quote}
sont \'equivalents, mais
\begin{quote}
\(<\){\it pomme/de/terre N;NDN:fs}\(>\) \(<\){\it cuire V:Kfs}\(>\)
\end{quote}
et
\begin{quote}
\(<\){\it pomme N:fs}\(>\) \(<\){\it de PREP}\(>\) \(<\){\it terre/cuite N;NA:fs}\(>\)
\end{quote}
ne le sont pas.

La grammaire accepte la s\'e\-quence si et seulement si on peut diviser la s\'e\-quence en portions qui sont chacune de l'un des deux types suivants~:
\begin{enumerate}
\item la portion est conforme \`a une s\'e\-quence de sortie \(v\) du transducteur~; la portion est \'equivalente \`a une s\'e\-quence grammaticale compl\`ete conforme \`a une s\'e\-quence d'en\-tr\'ee \(u\) associ\'ee \`a \(v\) par le transducteur. 
\item la portion est r\'eduite \`a une seule \'eti\-quette grammaticale compl\`ete, peu importe laquelle, mais le texte qui commence \`a ce mot n'admet aucun \'etiquetage grammatical qui corresponde \`a une s\'e\-quence d'en\-tr\'ee du transducteur.
\end{enumerate}
\begin{figure}
\centering
\includegraphics[width=7cm]{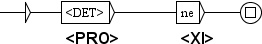}
\caption{le transducteur \(T_7\).}
\label{graphet7}
\end{figure}
Exemple~: le transducteur \(T_7\) (figure~\ref{graphet7}) accepte la s\'equence
\begin{quote}
\(<\){\it mais CNJC}\(>\) \(<\){\it aucun PRO:ms}\(>\) \(<\){\it ne XI}\(>\) \(<\){\it pouvoir V:P3s}\(>\) \(<\){\it d\'epasser V:W}\(>\) \(<\){\it ce DET:fs}\(>\) \(<\){\it limite N:fs}\(>\)
\end{quote}
qui est l'\'eti\-que\-tage correct de
\begin{quote}
{\it Mais aucun ne peut d\'epasser cette limite}
\end{quote}
(table~\ref{t7}). En revanche, si on remplace \(<\){\it aucun PRO:ms}\(>\) par \(<\){\it aucun DET:ms}\(>\) dans la s\'e\-quence grammaticale, on obtient une s\'e\-quence qui est rejet\'ee par \(T_7\).

\begin{table}
\centering
\begin{tabular}{|c|c|l|c|} \hline
S\'equences gramm. &	S\'equences d'en\-tr\'ee &	S\'equence gramm. &	S\'equences de sortie \\
compl. reconnues &	du transd. (\(u\)) &	compl\`ete &	du transd. (\(v\)) \\ \hline

-- &	-- &	\(<\){\it mais CNJC}\(>\) &	-- \\ \hline
\(<\)aucun DET:ms\(>\) &	\(<\)DET\(>\) &	\(<\){\it aucun PRO:ms}\(>\) &	\(<\)PRO\(>\) \\
\(<\)ne XI\(>\) &	ne &	\(<\){\it ne XI}\(>\) &	\(<\)XI\(>\) \\ \hline
-- &	-- &	\(<\){\it pouvoir V:P3s}\(>\)- &	-- \\ \hline
-- &	-- &	\(<\){\it d\'epasser V:W}\(>\) &	-- \\ \hline
-- &	-- &	\(<\){\it ce DET:fs}\(>\) &	-- \\ \hline
-- &	-- &	\(<\){\it limite N:fs}\(>\) &	-- \\ \hline
\end{tabular}
\caption{une s\'e\-quence grammaticale accept\'ee par le transducteur \(T_7\).}
\label{t7}
\end{table}

\begin{sloppypar}
\section{Conclusion}
\end{sloppypar}

Lorsqu'on examine l'\'eti\-que\-tage initial d'un texte tel qu'il est produit par {\tt INTEX}, comme dans la section 3.1, des id\'ees de r\`egles de le\-v\'ee d'am\-bi\-gu\"{\i}\-t\'es viennent spontan\'ement. Ces id\'ees se pr\'esentent parfois sous une forme telle que
\begin{quote}
Si le d\'eterminant {\it du} est suivi par un verbe, ce verbe ne peut \^etre qu'au participe pr\'esent,
\end{quote}
c'est-\`a-dire avec une condition qui reconna\^{\i}t une configuration grammaticale, et une contrainte grammaticale \`a imposer lorsque la condition est remplie. Il est facile de concevoir et de r\'ealiser une premi\`ere version d'une grammaire locale \`a partir d'une telle id\'ee. Les r\`egles que nous avons cit\'ees (sections 4 et 6) pour d\'ecider si une grammaire donn\'ee accepte une s\'e\-quence donn\'ee ne sont gu\`ere utiles \`a ce stade, car le fonctionnement du sys\-t\`eme est plus intuitif que ces r\`egles ne le laissent penser.

Toutefois, les intuitions grammaticales peuvent se r\'ev\'eler inexactes, \`a cause bien souvent d'une construction ou d'une ambigu\"{\i}t\'e \`a laquelle on n'aura pas pens\'e. Puisque nous nous sommes fix\'e l'objectif d'un taux de silence nul, les grammaires locales doivent \^etre test\'ees avec soin. C'est l\`a qu'il est n\'ecessaire de savoir en d\'etail ce que fera une grammaire une fois appliqu\'ee \`a des textes.

\begin{center}
\Large\bf
R\'ef\'erences
\end{center}
\parskip .6em
Brill, Eric. 1992. ''A Simple Rule-Based Part of Speech Tagger'', {\it 3rd Applied ACL}, Trente (Italie), pp.~152--155.

\noindent Courtois, Blandine. 1990. ''Un sys\-t\`eme de dictionnaires \'electroniques pour les mots simples du francais'', in {\it Langue fran\c{c}aise} 87, {\it Dictionnaires \'electroniques du fran\c{c}ais}, Paris~: Larousse, pp.~11--22.

\noindent Cutting, Doug, Julian Kupiec, Jan Pedersen, Penelope Sibun. 1992. ''A practical part-of-speech tagger'', {\it 3rd Applied ACL}, Trente (Italie), pp.~133--140.
\begin{sloppypar}
\noindent Dermatas, E., G.~Kokkinakis. 1989. ``A System for Automatic Text Labelling'', {\it Eurospeech 89}, pp.~382--385. 
\end{sloppypar}

\noindent DeRose, Stephen J. 1988. ``Grammatical Category Disambiguation by Statistical Optimization'', {\it Computational Linguistics}, vol.~14, no.~1, pp.~31--39. 

\noindent Federici, Stefano, Vito Pirrelli. 1992. ''A Bootstrapping Strategy for Lemmatization: Learning Through Examples'', {\it Papers in Computational Lexicography. COMPLEX 92}, F.~Kiefer, G.~Kiss, J.~Pajzs, \'eds., Institut de linguistique de l'Acad\'emie hongroise des sciences, Budapest, pp.~123--135.

\noindent Garside, Roger, Geoffrey Leech, Geoffrey Sampson. 1987. {\it The Computational Analysis of English}, Londres~: Longman.

\noindent Greene, Barbara, Gerald Rubin. 1971. {\it Automated Grammatical Tagging of English}, Rapport technique, D\'epartement de linguistique, Brown University, Providence, Rhode Island.

\noindent Hindle, Donald. 1983. ''Deterministic parsing of syntactic non-fluencies'', {\it 21st Annual Meeting of the Association for Computational Linguistics. Proceedings of the Conference}.

\noindent Hindle, Donald. 1989. ''Acquiring disambiguation rules from text'', {\it 27th Annual Meeting of the Association for Computational Linguistics. Proceedings of the Conference}, pp.~118--125.

\noindent Jelinek, F. 1985. ``Markov source modeling of text generation'', in {\it Impact of Processing Techniques on Communication}, J.K.~Skwirzinski, \'ed., Dordrecht.

\noindent Klein, S., R.F.~Simmons. 1963. ``A Computational Approach to Grammatical Coding of English Words'', {\it JACM} 10, pp.~334--347.

\noindent Knuth, Donald. 1973. {\it The Art of Computer Programming}, Addison-Wesley.

\noindent Koskenniemi, Kimmo. 1990. ''Finite-state parsing and disambiguation'', {\it Proceedings of COLING 90}, H.~Karlgren, \'ed., Universit\'e d'Helsinki, pp.~229--232.

\noindent Laporte, \'Eric. 1994. ''Experiments in lexical disambiguation using local grammars'',  {\it Papers in Computational Lexicography. COMPLEX 94}, Institut de linguistique de l'Acad\'emie hongroise des sciences, Budapest, 10~p.

\noindent Marshall, Ian. 1983. ``Choice of Grammatical Word-Class Without Global Syntactic Analysis: Tagging Words in the LOB Corpus'', {\it Computers in the Humanities} 17, pp.~139--150.

\noindent Milne, Robert. 1986. ''Resolving Lexical Ambiguity in a Deterministic Parser'', {\it Computational Linguistics}, vol.~12, no.~1, pp.~1--12.

\noindent Paulussen, Hans, Willy Martin. 1992. ''DILEMMA-2: a Lemmatizer-Tagger for Medical Abstracts'', {\it 3rd Applied ACL}, Trente (Italie), pp.~141--146.
\begin{sloppypar}
\noindent Pelillo, Marcello, Mario Refice. 1991. ''Syntactic disambiguation through relaxation processes'', {\it Eurospeech 91}, vol.~2, pp.~757--760.
\end{sloppypar}
\noindent Revuz, Dominique. 1991. {\it Dictionnaires et lexiques, m\'ethodes et algorithmes}, Th\`ese de doctorat, Publication~91-44 du LITP, Universit\'e Paris~7, 105~p.

\noindent Rimon, Mori, Jacky Herz. 1991. ''The recognition capacity of local syntactic constraints'', {\it 5th Conference of the European Chapter of the ACL. Proceedings of the Conference}, Berlin, pp.~155--160.

\noindent Roche, Emmanuel. 1992. ''Text disambiguation by finite-state automata, an algorithm and experiments on corpora'', in {\it COLING-92, Proceedings of the Conference}, Nantes.

\noindent Silberztein, Max. 1989. {\it Dictionnaires \'electroniques et reconnaissance lexicale automatique}, Th\`ese de doctorat, LADL, Universit\'e Paris~7, 176~p.

\noindent Silberztein, Max. 1990. ''Le dictionnaire \'electronique des mots compos\'es'', in {\it Langue fran\c{c}aise} 87, {\it Dictionnaires \'electroniques du francais}, Paris~: Larousse, pp.~71--83.

\noindent Silberztein, Max. 1993. {\it Dictionnaires \'electroniques et analyse automatique de textes. Le sys\-t\`eme INTEX}, Paris~: Masson, 233~p.

\noindent Vosse, Theo. 1992. ''Detecting and Correcting Morpho-syntactic Errors in Real Texts'', {\it 3rd Applied ACL}, Trente (Italie), pp.~111--118.

\end{document}